\documentclass{llncs}
\usepackage{times}
\usepackage{graphicx}
\usepackage{amsmath}

\DeclareGraphicsExtensions{.jpg, .pdf, .png}

\begin{document}

\title{Communicating and resolving entity references}

\author{R.V.Guha}

\institute{Google}

\maketitle

\begin{abstract}

Statements about entities occur everywhere, from newspapers and web
pages to structured databases. Correlating references to entities
across systems that use different identifiers or names for them is a
widespread problem. In this paper, we show how shared knowledge
between systems can be used to solve this problem. We present
"reference by description", a formal model for resolving
references. We provide some results on the conditions 
under which a randomly chosen entity in one system can, with high probability, be mapped 
to the same entity in a different system.

\end{abstract}

 \section{Introduction}

References to things/entities (people, places, events, products, etc.) are ubiquitous.
They occur in almost all communications, from natural language utterances to
structured data feeds. Correctly resolving these references is vital to the
proper functioning of many systems. Variations of this problem
have been studied in fields ranging from philosophy and linguistics to 
database integration and artificial intelligence. In this paper, 
we propose a framework for studying the reference problem.\\

One of the earliest descriptions of this problem was in Shannon's
seminal paper \cite{Shannon}. Shannon writes: ``The fundamental problem of
communication is that of reproducing at one point either exactly
or approximately a message selected at another point. Frequently
the messages have meaning; that is they refer to or are correlated
according to some system with certain physical or conceptual
entities.''. In other words, the symbols in a message are often
intended to refer to certain entities. The message can be said to be
fully understood only when the receiver can identify the intended
denotation of these symbols. \\

However, Shannon goes on to say:  
``These semantic aspects of communication are irrelevant
to the engineering problem.'' Nevertheless, he has given us most
of the tools to address this problem.  In particular, Shannon's 
model of communication is an excellent starting point for a framework for studying the 
problem of correctly resolving references to entities.\\

\subsection{Problem Model}

 In this paper, we use the model and terminology of communication theory.
  In classical information theory, the two parties agree on the set of possible 
messages. The sender picks one of these messages and transmits it through a channel. 
The communication is said to have succeeded if the receiver can correctly identify 
which of the possible messages was picked by the sender. \\

Similarly, in our case,  the two parties have to agree on the set of possible entities a
reference in message might refer 
to. In the most trivial case, if each entity has a unique identifier or name  (henceforth, 
simply referred to as name) and the two parties share the names for all the entities, 
assuming successful communication of the message, the receiver can trivially 
decode the intended entity references. \\

We are interested in the case where the sender and receiver do not share the 
names for all the entities. When the sender
and receiver don't share the name for an entity, the sender may be able to
construct a unique description of the entity only using terms that the
sender and receiver share names for. Such a  description can then
be used to refer to the entity.\\

For example, imagine communicating the identity of someone called  Michael Jones. 
Given the number of people with that name, the name alone is highly ambiguous. 
However, if we augment the name with the person's date of birth, his profession, etc., 
this description fairly quickly uniquely identifies the
person. \\

Disambiguating descriptions are ubiquitous in
natural language.  References to people, places and organizations in news articles are usually accompanied 
by short descriptions. By using symbols or names 
 whose meaning we share  and our shared view of the domain that we are 
 communicating about,  we construct descriptions that uniquely identify the entities 
 that we don't  share names for. 
Our goal is to formalize this mechanism so that it can be 
 used, in a reliable fashion, for communications between programs. \\

The sender refers to each of the entities that is in the message, that is not shared, by means 
of a unique description. It is the intent of the sender that only a single entity satisfies the 
description in the world visible to the receiver and that this entity be the one intended 
by the sender. When there is a difference in the view of the world as seen by the sender 
and receiver, it is possible that there are multiple entities satisfying the description or there 
is no entity satisfying the description. Even in such cases,  the receiver can guess at the intended 
referent of the description by selecting the entity that has the maximum likelihood of being 
the intended referent. In such cases, the sender, by augmenting the description with
additional information, can increase the likelihood of the receiver correctly interpreting
the description. This is analogous to using coding to overcome noise
in the channel. \\

\subsection{Summary of Results}
There are many interesting questions that can be formulated in
our framework. We list some of them here, along with informal
descriptions of the results presented in the rest of the paper.

\begin{enumerate}
\item The minimum number of names that need to be shared for the 
sender to successfully communicate references to all other entities: 
Our most interesting result is that the amount of shared knowledge 
required to decode the intended denotations of the terms in a message 
is inversely proportional to the channel capacity required to transmit
the message.\\

\item The minimum length/information content of the description. We
  find that the average length of the description required is
  inversely proportion to the channel capacity required.
  This is closely related to the minimum number of names that need to be shared.\\

\item Various classes of descriptions and their sharing
  requirements. We find that as we allow for more complex descriptions
  that are computationally more difficult to decode, the amount that
  needs to be shared decreases. In other words, in analogy with the
   space vs time tradeoff typically found in computation, we find a
   time vs sharing required tradeoff in communicating references.\\

\item The communication overhead of communicating references: Bootstrapping
  from the minimum number of shared names (versus sharing all names)
  incurs both computation and communication overhead. We find that the
  communication overhead is independent of the entropy of the
  underlying world. Worlds with higher entropy are more difficult to
  compress, but require shorter descriptions and vice
  versa. Interestingly, these two effects cancel out, giving us a
  constant overhead which is purely a function of the description
  language. 
\end{enumerate}

\section{Outline of paper}

 We first present our model of correlating or communicating references as an extension
 of Shannon's model of communication. We then review prior work in
 terms of this framework. We then formalize the concept of descriptions and the entropy
 of these descriptions. Finally we provide some results on the conditions under
 which the communication can take place.

\section{Communication model}
 In this section, we describe our extended model of communication. We start with 
the traditional information theory model in which the sender picks one of a possible set of 
messages, encodes it and transmits one of these through a potentially noisy 
channel to a receiver. We add the following to this model.

\begin{enumerate}
\item There is an underlying 'world' that the messages are about. Our model of
the world has to be expressive enough to represent most likely domains of discourse.
A wide range of fields, from databases and artificial intelligence to number theory
have modeled the world  as a set of entities and a set of N-tuples on these 
entities. We use this model to represent the underlying world. 
Since arbitrary N-tuples can be constructed out of 3-tuples, we can restrict
ourselves to 3-tuples, which is equivalent to a directed labelled graph.
We will henceforth refer to the world that the communication is about, as 'the graph', 
the N-tuples as arc labels and the entities as nodes. Without loss of
generality, we assume that there is at most a single arc between any
two nodes.\footnote{Given a graph which allows multiple directed arcs between any pair of nodes, we map it to a 
corresponding graph which has at most one undirected arc between any pair of nodes, with the 
same set of nodes, but different set of arc labels. The set of arc labels in this reduced 
graph are the different possible combinations of arc labels and arc directions in the 
original graph that may  occur between any pair of nodes. So, given N arc labels 
in the original graph, we might  have upto $2^{N+1}$ labels in the reduced graph. 
} We represent the graph by its adjacency matrix. The entries in the adjacency
matrix are arc labels. If there is an arc with the label $L$ between the nodes
$V_1$ and the node $V_2$, the cell in the adjacency matrix in row/column $V_1$, row/column $V_2$ will
have the entry $L$. We will use the syntax $L(V_1, V_2)$ 
to say that there is an arc labelled $L$ from $V_1$ to $V_2$.
\\

 The sender and receiver each see a subset of this graph. We
  consider both the case where their view of the graph is the same and
  where their views of the graph differ. In the second case, they
  might have visibility into different parts of the graph and/or there
  may be differences in their views of the same portion of the
  graph.  We are not interested in  which is the correct view, but merely in how
in the overlap between the two views affects communication. \\

\item Nodes in the graph may be entities (people, places, etc.) or
  literal values such as strings, numbers, etc. \\

\item Each entity and arc label in the graph has a unique name. Some subset 
of these names are shared by the receiver and sender. In particular, 
all the arc labels are shared. Literals (numbers, strings, etc.),
since they don't have any identity beyond their encoding, are assumed
to be shared.\\

\item Each message encodes a subset of the graph. The communication is said to be successful if the receiver correctly 
identifies the nodes in the graph contained in the message. There may be arcs 
in the message that are not in the receiver's view of the world. 
These could be the content of the message.

\end{enumerate}

\subsection{Simplifying Assumptions}
 We make the following simplifying assumptions for our analysis.
\begin{enumerate}
  \item We will assume that the sender and receiver share the grammar 
          with which the graph is encoded. The details of the grammar 
          are not relevant, so long as the receiver can parse the message. \\

   \item The graphs transmitted can be expressed a set of source, arc-label, target triples. 
I.e., no quantifiers. Disjunctions and negations are in principle allowed. 
We map these into corresponding simple triples without these connectives, on a different graph.

\end{enumerate}

\subsection{Examples}

We look at a few examples of our model of communication and the use of descriptions
to refer to entities.In all these examples, the graphs have a single arc label (call
it P).

\begin{enumerate}

\item In the example shown in Fig 1, both parties observe the same
  graph. The names for the nodes B and D are shared.
The sender sends the sequence of symbols  ``$P(Q, T)$''. Given the underlying graph, since Q and T are known to be 
not B or D, the receiver can map Q to either S (which would be
correct) or to R (which would be wrong). The sender understands the
potential for this confusion and adds the description ``$P(B, Q)$'' to the
message, eliminating the wrong mapping as a possibility.
If either B or D were not shared, there will be at least one node
whose reference cannot be communicated. \\

\begin{figure}
\centering
\includegraphics[width=4.0in]{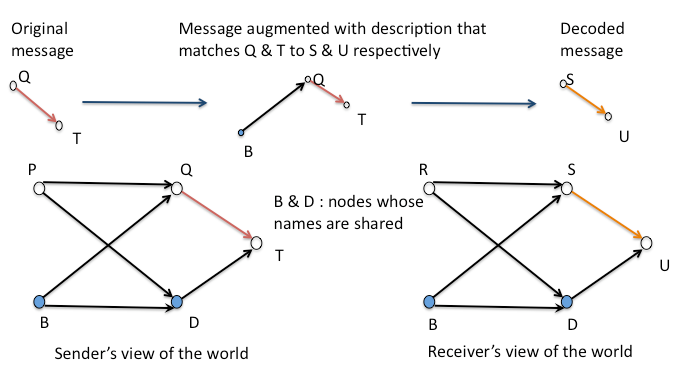}
\caption{Distinguishing descriptions with shared nodes}
\end{figure}

\item In the example shown in Fig 2, the underlying graph is slightly
richer than the graph in Fig 1. Because of this additional richness,
even without the names of any of the nodes being shared, the sender
can construct distinguishing descriptions for all the nodes in the graph.
However, the size of these descriptions is much bigger. \\

\begin{figure}
\centering
\includegraphics[width=4.0in]{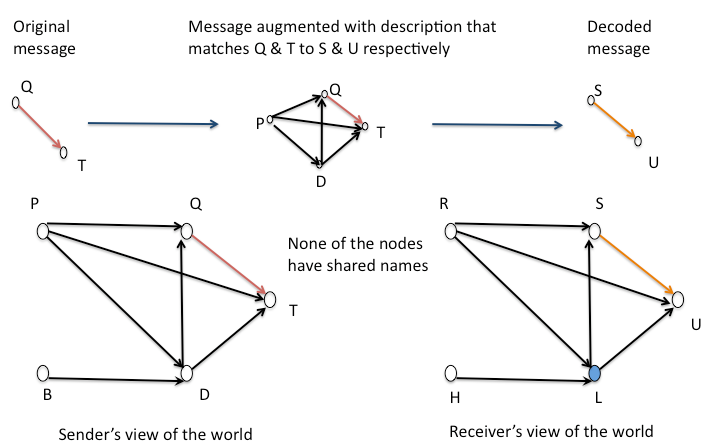}
\caption{Distinguishing descriptions without shared nodes}
\end{figure}

 \item In the example in Fig. 3, The underlying graph is a clique. 
In this case, none of the nodes have 
descriptions that distinguish them from any of the other nodes. In order
to communicate a reference to a node, the sender and receiver have to
share its name. \\

\begin{figure}
\centering
\includegraphics[width=2.0in]{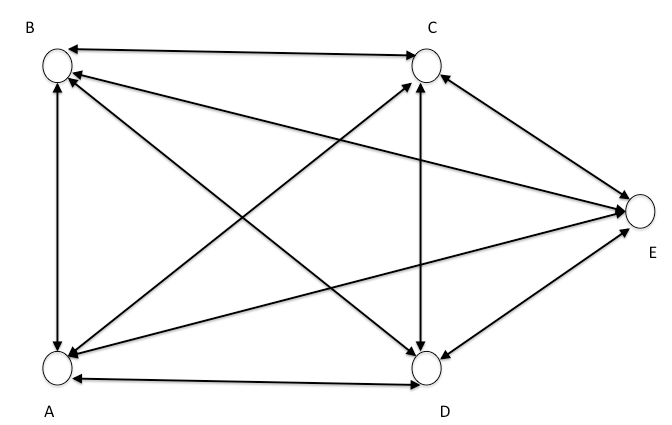}
\caption{Graph with no distinguishing descriptions}
\end{figure}

 \item In the example in Fig. 4,the sender and receiver have different views of the underlying
   graph. This difference causes the distinguishing description to
   be wrongly interpreted, leading to the receiver incorrectly interpreting
   the intended reference.

\begin{figure}
\centering
\includegraphics[width=4.0in]{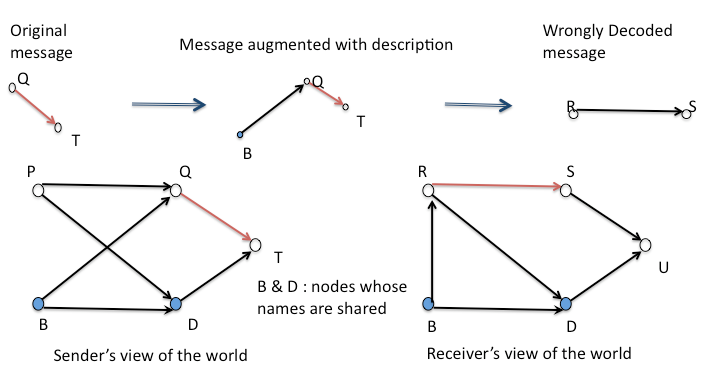}
\caption{Wrong communication due to different views}
\end{figure}

\item In the example shown in Fig 5, the sender adds redundant
descriptions to the nodes in the message. Even though there is no node
on the receiver's side that satisfies the entire description, only the
correct mapping satisfies the maximum number of literals in the
description. This illustrates how the sender and receiver can
communicate even when they don't share the same view of the world. 
As with communication on a noisy channel, by using slightly longer 
messages, the sender can, with high probability, communicate the intended references.

\begin{figure}
\centering
\includegraphics[width=4.0in]{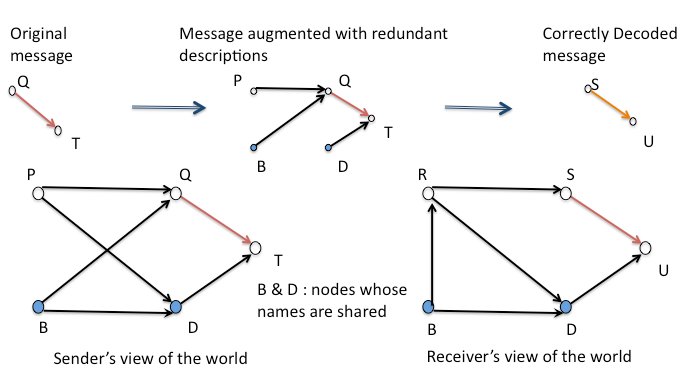}
\caption{Correct communication despite different views}
\end{figure}
\end{enumerate}

\section{Related Work}

The problem of correlating references to entities across systems arises
in many different fields, including statistics, epidemiology, history, census analysis,
database integration, privacy protection, linguistics and communication.  \\

Most of the work that has been done on this problem has been in computer science, though
Shannon was the first to identify the problem in its most general form.
Even in computer science, the problem goes under many different names, including 
"record linkage", "list washing", "merge/purge processing", "data matching", "entity disambiguation", 
"coreference resolution" and "database hardening". \\

We now show how some well known entity resolution problems can be mapped into this
framework.

\begin{itemize}

\item {\bf Data Integration}: One common model that shows up in database integration, processing
  of catalogs (e.g., product catalog merging), record linkage, etc. is
  as follows. We are given a list of items (e.g., people, products.) each with a set of literal valued attributes 
(e.g., name, address, age, price, phone number). The values of these attributes may be noisy, 
with errors introduced by typos, alternate punctuation, transcription errors, etc. \\

 In record linkage, we have a bag of such items, wherein multiple
 items might correspond to the same entity. The goal is to 'link'
 these duplicate records. \\

 In catalog/feed processing, there is a master
 database of entities and we are given a new set of entities, each
 with some attributes. Some of the new entities may correspond to
 existing entities. The goal is merge the new data into the master
 database, correctly identifying entities which already exist in the 
 database. \\

This class of problems maps into our framework as follows. In the case of record
linkage, the transmitter is the given record (for which we are trying
to find a duplicate) and the reciever is the rest of the database. In
the case of catalog processing, the new data is the transmitter and
the existing database is the reciever. \\

The underlying world is modelled as a bipartite graph with entities on one
side and attribute values on the other side.
The attribute values, being literals, are assumed to be shared. There
are two sources of problems. Sometimes, there isn't enough information to
conclude that two items correspond to the same entity (even if all
their attributes are the same). For example, if we have two items and
all we knew about each item was that it has the name 'Michael Jones',
we cannot conclude one way or the other whether the two entities are
the same. Often, there are differences in the values of the attributes
which can lead to problems. For example, the entity in the database
might have the name 'Michael Jones' whereas the item in the feed might
have the name 'Mike Jones'. Both of these case are handled in our
framework in terms of the difference in the world view between the
reciever and transmitter. The relationship between attribute values
like 'Michael Jones' and 'Mike Jones' is captured by the mutual
information between the two world views. \\

 The research by (\cite{Elmagarmid}, \cite{Cohen} and \cite{Pasula}) are archetypal of the approaches 
 that have been followed for solving this class of problems.
 Because of the simplicity of the model, much of the attention has focussed on the development of algorithms capable of correctly 
 performing the matching between attributes. Further, most of the work has focussed on overcoming the lexical 
 heterogeneity of the representation of the string values and on differences introduced by 
data acquisition and entry errors. \\

The work presented here differs in two main respects. Firstly, the data/representation model used to  
encode information about each entity is more expressive, allowing for arbitrary relational information. 
The methods proposed in the research on data integration typically do
not extend to complex relational structures.  
Secondly, our goal is not to come up with a specific matching algorithm, but to establish a general
 framework and derive bounds on the knowledge that must be shared and
 for the minimum length/information content of the description for the matching to be possible at all.  \\

\item {\bf Privacy}: Information sharing, while essential for many
  transactions, leads to loss of privacy. Often, we would like to
  determine how much information can be shared about an entity without
  uniquely identifying it. We map this into our formalism as
  follows. The information that is being revealed is the transmitters
  view of the world. The reciever of the information is the same as
  the reciever in our model. Our model also allows for the transmitter
  to understand the impact of modifying pieces of information that are
  not essential to the transaction, which might help preserve the
  privacy of the entity that information is being shared about.\\

\item {\bf Language Understanding}: Pronoun and anaphora resolution is
  one of the big problems in language understanding. Though the hard
  problem is that of going from the natural language to a formal
  representation (such as a graph), once that is done, the task of
  going from a pronoun or anaphora can be understood in terms of our
  framework. The receiver's world view consists of the set of
  candidate references to the pronoun/anaphora and facts known
  about them. The transmitter's world view is the facts known about
  the pronoun/anaphora.

\end{itemize}

\section{Graph model}
  As we saw in  examples 1, 2 and 3, graphs differ in their ability to support distinguishing 
descriptions. The identity of nodes can be communicated by the uniqueness of the shape of
 the graph around them and by their relation to one or more shared nodes. If the structure
around every node looks like the structure around every other node, it becomes more difficult 
to construct unique descriptions. As the richness of the graph increases,
 the number of candidate unique descriptions for a given set of shared names increases. 
The entropy of the graph is a measure of its richness.  \\

  We first need a mathematical model for our graph.  We assume that our 
graph is created by a stochastic process. There has been extensive work on 
modeling graphs created by stochastic processes, most of 
which can be easily extended to  labelled graphs. We begin with a set of $N$ vertices 
and then add edges between pairs of vertices according to some probability distribution. 
Different probability distributions give us graphs with different kinds of properties. 
The most studied is the Erdos Renyi model, denoted $G(N,p)$, in which we have 
a graph with $N$ nodes and 
every possible edge occurs independently  with probability p. In the labelled graph 
variant of this model, we have a probability distribution where the the probability of the arc 
between any pair nodes having the label $L_i$ is $p_i$, with the absence of any 
arc being considered a special arc which we shall refer to as $L_{null}$. \\

Many other models have been proposed for random graphs. 
Recently here has been considerable work on other random graph models \cite{newman}, 
such as those involving preferential attachment, which can be useful for modelling 
structures such as the web.  Some systems use more 'regular' graphs (a grid
being an extreme example of such a regular graph). Database systems with strict schemas
are a good example of this. The choice of graph model depends on the details of 
the underlying world that the sender and receiver are exchanging  messages about. \\

The analysis presented in this paper can be used with any of these  models. 
Our only requirement is that 
certain rows in the adjacency matrix should be generated by an ergodic
process, which basically means that different randomly chosen long enough substrings
 from these rows in the adjacency matrix should have the same distribution of arc labels. 
More concretely, randomly 
chosen long enough samples from these rows in the adjacency matrix should obey the 
asymptotic equipartition property (AEP) \cite{Cover}. The AEP states that if we 
have a process generating strings of length K according to a probability distribution 
that has an entropy H, the set of $2^K$ possible strings can be partitioned into two sets: 
the first set of size $2^{HK}$, which is called the typical set, of strings that are 
likely to occur, and the second set, containing the remaining strings, that are not 
likely to occur. Each of the strings in the typical set are have an equal probability 
of occuring, which is $2^{-HK}$.

\section{Shared knowledge}

   Uniquely identifying descriptions work because of shared knowledge. 
When the sender describes a node X as $L_1(X, S_1)$, i.e., by specifying 
that there is an arc labelled $L_1$ between X and the shared node $S_1$, she 
expects the receiver to know both the shared name for the node $S_1$ and 
to know which nodes have arcs labelled $L_1$ going to $S_1$. If either 
of these two conditions is not met, the description will not serve its purpose. 
We distinguish between the two  kinds of shared knowledge: shared names and shared 
knowledge of the graph.

\subsection{Sharing Names}
  We are interested in determining the minimum number of nodes whose names 
need to be shared.  We  assume that the names for the arc labels are shared. 
We are interested in the case where the structure is very large and there are a 
small fixed number of arc labels, so that the number of arc labels is very small 
compared to the number of nodes. In such cases, assuming that the arc labels 
are shared should have a very small effect. The quantitative measure of sharing 
is very simple --- it is simply the number of nodes whose names are shared.

\subsection{Shared knowledge of the graph}

Quantifying the sharing of graph is more subtle than quantifying the sharing of names. What is shared
 is as important as how much is shared. Differences in the views of the sender and receiver 
change the effective graph entropy that descriptions can exploit. For example, if 'color' 
is one of the attributes of nodes and the receiver is blind, then the usable entropy of the 
graph, i.e., the number of candidate descriptions, is reduced. On the other hand, if the
 receiver is  color blind, some of the values of color (such as black and white) 
may be correctly recognized while there may be limited ambiguity in other values 
such as red or green. We use the mutual information between the sender's and receiver's 
versions of the graph's adjacency matrix as the measure of how much knowledge of 
the underlying world is shared.

\begin{eqnarray*}
   M &=& H(Sender) - H(Sender|Receiver) \\
          &=& H(Receiver) - H(Receiver|Sender) 
\end{eqnarray*}

\section{Descriptions}
  A description of a node is any subgraph of the graph, which includes that 
node and some (possibly none) of the nodes whose names are shared. Since any 
subgraph that includes  a node is a description of that node, every node will 
have many descriptions. Some of these descriptions may uniquely identify the node. \\

 Descriptions come in many different 'shapes'. The computational complexity of
dereferencing a description is a function of its shape. If a description is an
arbitrary subgraph, dereferencing it requires the receiver to solve a subgraph
isomorphism problem, which is known to be NP-complete. However, if we impose 
some restrictions on the structure of admissible descriptions, the complexity
of decoding the description can be kept down. In this section, we look at a 
few different kinds of descriptions with different levels of decoding complexity.\\

Assume that the sender and receiver share names for a set 
of K nodes {$ S_1, S_2, ... S_K$}. We have $M$ arc labels: $<L_1, L_2, ...L_m>$.  
Given a node $X$ (whose name is not shared), we need to  construct a description  
for this node. Let the relation
between this node and the $i^{th}$ of the $K$ nodes be $L_{xi}$.  The relation 
could be a direct arc between the two nodes or a more complex path.
The simplest class of descriptions, which we will refer to as 'flat descriptions', 
corresponds to the logical formula:

\begin{equation*}
L_{x1}(X, S_1) \wedge L_{x2}(X, S_2) \wedge ... \wedge L_{xK}(X,S_K)
\end{equation*}

In this class of descriptions, if there is no direct arc between $X$ and the shared 
node $S_i$, we use the special arc label $L_{null}$. This class of  descriptions can 
be decoded very efficiently, using standard database techniques.\\

 We can also write this as the string $L_{x1}L_{x2}L_{x3}...L_{xK}$. If the columns 
corresponding to the $K$ nodes whose names are shared are placed adjacent to each 
other in the adjacency matrix, this string is simply the entries in those columns for the 
row corresponding to $X$ in the adjacency matrix. As mentioned earlier, the only assumption we make about the graph is that these description
strings (i.e., the rows/columns of the adjacency matrix corresponding to the K shared terms)
obey the AEP. The entropy of this class of description strings is simply

\begin{equation*}
                             H_d = - \Sigma p_i log (p_i) 
\end{equation*}

where $p_i$ is the probability of the label $L_i$ occuring between two
randomly chosen nodes in the graph.

More complex descriptions emerge when, instead of using $L_{null}$ for the case where there is
no direct arc between $X$ and $S_i$, we allow paths or of length longer than 1.
More generally, we can allow arbitrary intermediate subgraphs connecting $X$ and $S_i$,
involving multiple intermediate nodes with arcs between these intermediate nodes. 
Depending on the class of intermediate subgraphs allowed, we get different kinds 
of descriptions with different levels of dereferencing complexity. In increasing
order of complexity, we can restrict ourselves to strict paths, trees, planar intermediate 
subgraphs or allow for arbitrary intermediate subgraphs. As the complexity of 
the allowed intermediate graph increases, the number of possibles
shapes for the graph and hence the entropy of the descriptions increases.\\

In this paper, we restrict our analysis to descriptions where the intermediate graph
is of some fixed size $D$. 
Let us name the set of possible  graphs of size $D$ with an arc label set $<L_1, L_2, ...L_m>$ as
$<L_{null}, L_{D1}, L_{D2}, ...>$. If $D = 1$, then this set is just $<L_{null}, L_1, L_2, ...L_m>$. 
When $D > 1$, the description for $X$ looks the same as when $D > 1$, except, when there is no direct arc between
$X$ and $S_i$, we check to see if there is an intermediate graph of size $\leq D$
connecting $X$ and $S_i$ and if there is, we use the corresponding name for it.
Let the entropy of this description string be $H_D$. 
Consider a transformation of the adjacency matrix where the  $L_{null}$s are 
replaced with the appropriate terms from $<L_{null}, L_{D1}, L_{D2}, ...>$. 
$H_D$ is the entropy of strings from this adjacency matrix and $M_D$ is the mutual information 
between the sender's and receiver's views of this adjacency matrix.\\

Since there may be multiple, non-isomorphic intermediate graphs of size $D$ between 
$X$ and $S_i$, to identify a unique $L_{Dj}$ that can be used as the entry for
the appropriate cell in the adjacency matrix for the arc between $X$ and $S_i$, the
sender and receiver can establish a total order over the set of possible graphs of
size $\leq D$ and use the most preferred graph that occurs between $X$ and $S_i$. For 
computational reasons, the total order should prefer smaller graphs, but the analysis
of is independent of which graph is preferred.

\subsection{Entropy of complex descriptions}
 
 Since the set of possible replacement values for $L_{null}$ increases as the richness
of the possible intermediate graph grows, the entropy of the description string
also grows with the richness of descriptions. We are interested in the growth
of the entropy of descriptions ($H_d$) where the number of nodes in an intermediate
description is $D$ as a function of $D$ and entropy of the graph $H_g$. For the sake
of this analysis, we will ignore automorphisms.\\

 Each possible intermediate graph of size $D$ is a sub-block (of potentially non-contiguous 
rows and columns) of the adjacency matrix of the graph that is $D$ columns wide and $D$
rows tall. Even though there are $2^{D^2}$ possible graphs (ignoring automorphisms)
of size $D$, as per the AEP, only $2^{H_gD^2}$ are likely to occur and each occurs
with a probability of $2^{-H_gD^2}$. We put these graphs in a total order and name them
$<L_{D1}, L_{D2}, ...>$ so that if both $L_{Dj}$ and $L_{Dj+l}$ occur between $X$ and $S_i$,
we ignore the latter. The probability of the subgraph $L_{Dj}$ occurring between two random nodes is:

\begin{equation*}
P(L_{Dj}) = 1 - (1 -2^{-H_gD^2})^{N \choose D}
\end{equation*}

Since this is the same for all $j$, we will write this simply as $P(L_D)$. 
The probability of a particular cell in the adjacency matrix
revised for descriptions of size $D$ containing $L_{Dj}$ is the probability of not
having any intermediate graph that is preferred over $L_{Dj}$ between $X$ and $S_i$ and
$L_{Dj}$ occurring between $X$ and $S_i$, i.e.,

\begin{equation*}
PA(L_{Dj}) = P(L_D)(1 - P(L_D))^{(j-1)}
\end{equation*}

and the entropy of the descriptions is

\begin{equation*}
H_D = \sum_{i=0}^{2^{H_gD^2}} -PA(L_{Di})log(PA(L_{Di}))
\end{equation*}

For the special case where  D is equal to N, $P(L_D) = 2^{-H_gN^2}$.
If we prefer the biggest intermediate graphs, ignoring automorphisms, the entropy $H_D$ is

\begin{equation}
H_D = \sum_{i=0}^{2^{H_gN^2}} 2^{-H_gN^2} log(2^{-H_gN^2}) = H_gN^2
\end{equation}

\section{Minimum Sharing Required}

In this section, we compute the minimum number of nodes that need to be shared as
a function of the entropy of (probability distribution associated with) the 
descriptions, i.e., $H_D$.
Assume that the sender and receiver share names for a set 
of K nodes ${S_1, S_2, ...}$ and let the description string for $X$ be 
$L_{x1}L_{x2}L_{x3}...L_{xK}$. The K nodes are selected such that $H_D$ is maximized. In
the case of an Erdos Renyi random graph, we can choose any random set of K nodes. For
other graphs, the descriptions associated with different sets of K nodes will have
different entropies.\\

When the receiver gets the description  $L_{x1}L_{x2}L_{x3}...L_{xK}$, she can 
easily deference it by looking up the nodes that are in the relation $L_{x1}$ 
with $S_1$ and $L_{x2}$ with $K_2$, etc. Depending on the size of the description, 
we may end up with more than one such node. We would like to determine the minimum 
value for K, which would also be the minimum number of nodes for which names need to 
be shared, so that, with high probability, we have only one node that dereferences 
to the description.

\subsection{Case 1: Identical views of the graph}

 \textbf{Theorem:} Let the sender and receiver share names for $Glog(N)/H_D$ nodes, 
where $H_D$ is the entropy of the descriptions used by the sender to identify entities 
and $N$ is the number of nodes in the graph. For large graphs, if $G \geq 2$ then, 
with high probability, the sender and receiver can communicate references to all 
but a constant number of the other nodes. If $G <  2$, then, with high probability, 
there will be more than a constant number of nodes that the sender and receiver cannot 
communicate references to. \\

\textbf{Proof:}

 Let the entropy of the string $L_{x1}L_{x2}L_{x3}...L_{xK}$ is $H_D$.
According to the Asymptotic Equipartition Property, the set of possible descriptions 
of length K can be partitioned into 2 sets, one of site $2^{KH_D}$ descriptions, the 
'typical set', with probability of each description in this set being $2^{-KH_D}$ and 
the other set containing the rest of the descriptions, which have a negligible 
likelihood of occurring. Each of the $N$ objects in the graph has a description that 
comes from the typical set. Since the likelihood of each of these descriptions is equal, 
we can model the $N$ descriptions as coming from a random sampling with replacement 
of the typical set. As $K$ increases, the number of candidate descriptions (i.e., the 
typical set) increases. We want to compute the smallest $K$ so that the expected number of 
distinct samples from $N-K$ selections with replacement (which would be the descriptions 
for the nodes whose names are not shared) is $\approx N-K$. If $K << N$, we can approximate this 
with smallest $K$ so that the expected number of distinct samples from $N$ selections with 
replacement is sufficiently close to $N$. We would like the smallest value of $K$ such 
that the expected number of unique descriptions is at most a small constant (say $C$) away from $N$. \\

 This problem is special case of a well studied problem that appears in the birthday paradox, 
occupancy problem, collision estimation in hashing, etc. In the occupancy problem, we have $N$ 
balls that are randomly distributed across $J$ bins. In the birthday paradox, we have $N$ people 
in a room and we are interested in the likelihood of two or more of them having the same birthday. 
In the hash collision problem, we have $N$ items being hashed into $J$ hash buckets and are interested 
in the estimated number of hash collisions. In our case, each ball/item corresponds to a node and 
each bin/bucket corresponds to a candidate description. We are interested in determining how many 
candidate descriptions we need so that a random allocation of nodes across these descriptions 
leaves at most a constant number of nodes with more than one description (these nodes can then 
be shared, and if there are only a constant number of them, as $N$ grows large, we can ignore these). 
From \cite{cormen} we know that the estimated number of collisions is $C = N^2/2J$. $J$, 
the number of descriptions, is equal to $2^{H_DK}$. Substituting, we get,

\begin{equation*}
C2^{H_DK+1} = N^2
\end{equation*}
\begin{equation*}
log(C) + 1 + H_DK = 2log(N) \\
\end{equation*}

Ignoring (1 + log(C)) for large N, we get,

\begin{equation*}
 K \approx 2log(N)/H_D
\end{equation*}

 This shows that $2log(N)/H_D$ is an upper bound on the number of nodes that need to be shared.  \\

Now, we show that this is also a lower bound. Let $J = N^G$. In this case, $K = Glog(N)/H_D$. 
The number of collisions, $C = N^(2-G)$. Clearly, if $G \neq 2$, the number of collisions 
grows as $N$ grows. Hence, $2log(N)/H_D$ is a lower bound as well.

\subsection{Different views of the graph}

 It is possible for the sender and receiver to correctly communicate
 references even when there are differences between their views of the
 underlying graph.  We use the mutual
 information ($M_D$) between the two graphs (the one seen by the
as the measure of the shared
 knowledge. Our proof is very similar to the proof for the Shannon's
 theorem.  \\
 
\textbf{Theorem:} Let the sender and receiver share names for $Glog(N)/M_D$ nodes, where $M_D$ is 
the mutual information between the views of the graph that the sender and receiver are communicating 
about and $N$ is the number of nodes in the graph. For large graphs, if $G \geq 2$ then, with high 
probability, the sender and receiver can communicate references to all but a constant number of the 
other nodes. If $G <  2$, then, with high probability, there will be more than a constant number of 
nodes that the sender and receiver cannot communicate references to. \\

\textbf{Proof:} As before, the receiver and  sender share the names for K nodes. Receiver gets 
the description $ L_{x1}L_{x2}L_{x3}...L_{xk}$. Either many, exactly one or zero nodes in the receiver's 
graph match this description. The receiver looks at each object in his side and  considers the set of $K$ 
long descriptions that could be on the sender's side for that node. This set is of size $2 ^ {(K H_{(S|R)}}$, 
where $H_{(S|R)}$ is the conditional entropy of the sender's description, given the receiver's description. 
There are $2^{KH_D}$ descriptions of length K on the sender's side. So, there are $2^{K (H_D - H_{(S|R)}}$ = $2^{KM_D}$ sets of descriptions of size  $2 ^ {(KH_{(S|R))}}$, i.e., $2^{KM_D}$ 'mapping sets' on the sender's side.   \\

There are N nodes, which randomly pick amongst these $2^{KM_D}$ mapping sets. If $K << N$, we want to compute the smallest $K$ so that the expected number of distinct samples from $N$ selections with replacement is sufficiently close to $N$. Ideally, we would like a bound on $K$ so that the expected number of unique descriptions is at most a small constant (say C) away from $N$. This is exactly the same problem we solved before. Using the same approach, we get  \\

\begin{equation}
    K \approx 2\frac{log(N)}{M_D} 
\end{equation}

\subsection{Description Length}
 
 For the case where the sender and reciever have the same view of the
 world, it follows from the proof of the earlier theorem that the
 description has to be at least $2log(N)/H_D$ long. The information
 content of the description has to be at least $2log(N)$. For the case
 where there is a difference in the views, the description has to be
 of length $2log(N)/M_D$ and the information content of the
 description has to be at least $2log(N)H_D/M_D$. 

In cases (such as with record linkage and catalog merging), where the
graph reduces to a set of entities with literal attribute values,
since the literals are shared, there is no dearth of shared symbols.
The description length/information content can be used to determine
whether we have enough information about an entity to map it to some
other entity. We can also use it to determine how much information we
can reveal about someone without revealing their identity.

\subsection{Discussion}

\begin{enumerate}

\item The number of nodes that need to be shared is inversely
  proportion to the entropy and hence the channel capacity required to send the message. Messages
  that are  more compressible need more shared names to correctly
  resolve all entity references. In the extreme, for a clique which has
  zero entropy, every name needs to be shared.  \\

\item As the richness of the description language grows, $H_D$ grows
and the minimum number of nodes that need to be shared reduces. 
In the limit, if $H_D = 2log(N)$, only one name needs to be shared. 
If $D = N$, then $H_D = H_gN^2$. So, if $H_g > 0$, if the receiver 
is able to decode sufficiently large graphs, we don't need more 
than a constant number of nodes with shared names.\\

\item We can look at this as an addressing problem: With an optimal use of log(N) 'bits', 
we can construct unique addresses for N items. However, use of the address space is less 
than optimal in two ways. First the entropy of the graph $H_D$ tells us how efficiently 
each 'bit' is used. Second, we loose a factor of 2 because of the random allocation 
of addresses to items.\\

\item As the number of shared nodes increases, the required entropy
     decreases and the complexity of decoding descriptions decreases. \\

\end{enumerate}

\subsection{Communication overhead}

 In this section, we consider the communication overhead of using  descriptions. 
Consider the overhead in sending a single triple containing 2  nodes whose names are not shared.\\

In addition to the triple itself, we have 2 descriptions, each of
which is of size $2log(M)log(N)/H_D$ where $M$ is the set of possible
entries in the adjacency matrix, in the class of descriptions admitted (i.e., the vocabulary or number of alphabets in the description language).
Since the descriptions themselves are strings from the adjacency matrix, they can be compressed during
communication. Since their entropy is $H_D$, their size after
compression is each $2log(M)log(N)$ and the total overhead is $4log(M)log(N)$.\\

If the names of the 2 nodes were shared, we would need $2log(N)$ bits to
express  the  names of the 2 nodes. So, the overhead of using descriptions
instead of names is a factor of $2log(M)$.\\

As the richness of the description language grows, $log(M)$ increases
and the communication overhead increases, computational complexity increases
and sharing requirement decreases. \\

Very interestingly, for a given vocabulary of descriptions, the communication overhead 
is independent of the entropy of the description language and the number of nodes 
whose names are shared.

\section{Acknowledgements}

 I would like to thank Andrew Tomkins and Phokion Kolaitis for providing me a home at IBM research to start this work. I would also like to thank Bill Coughran for encouragement to finish this at Google. Finally, I thank Vineet Gupta, Andrew Moore and Andrew Tomkins for feedback on drafts of this paper.

\bibliographystyle{abbrv}
\bibliography{cr}

\end{document}